\documentclass[conference]{IEEEtran}
\usepackage{stmaryrd}
\usepackage{amsfonts}

\usepackage{graphicx,times,epsfig,amsmath} 
\usepackage{balance,amssymb}
\usepackage[mathscr]{eucal}
\usepackage[square,sort,comma,numbers]{natbib}
\usepackage{url}

\usepackage[breaklinks]{hyperref}
\usepackage{subfig}
\usepackage{float}
\usepackage{algorithm}
\usepackage{algorithmic}

\graphicspath{{figs/},{figs/faces/}}

\IEEEoverridecommandlockouts    

\begin{document}

\title{\ \\ \LARGE\bf Overfitting Mechanism and Avoidance in Deep Neural Networks\thanks{Shaeke Salman and Xiuwen Liu are with the Department of Computer Science, Florida State University, Tallahassee, Florida 32306, USA (email: \{liux,salman\}@cs.fsu.edu).}}

\author{Shaeke Salman and Xiuwen Liu}


\maketitle

\begin{abstract}
  Assisted by the availability of data and high performance computing, deep learning techniques have achieved breakthroughs and surpassed human performance empirically in difficult tasks, including object recognition, speech recognition, and natural language processing. As they are being used in critical applications, understanding underlying mechanisms for their successes and limitations is imperative. In this paper, we show that overfitting, one of the fundamental issues in deep neural networks, is due to continuous gradient updating and scale sensitiveness of cross entropy loss. By separating samples into correctly and incorrectly classified ones, we show that they behave very differently, where the loss decreases in the correct ones and increases in the incorrect ones. Furthermore, by analyzing dynamics during training, we propose a consensus-based classification algorithm that enables us to avoid overfitting and significantly improve the classification accuracy especially when the number of training samples is limited. As each trained neural network depends on extrinsic factors such as initial values as well as training data, requiring consensus among multiple models reduces extrinsic factors substantially; for statistically independent models, the reduction is exponential. Compared to ensemble algorithms, the proposed algorithm avoids overgeneralization by not classifying ambiguous inputs. Systematic experimental results demonstrate the effectiveness of the proposed algorithm. For example, using only 1000 training samples from MNIST dataset, the proposed algorithm achieves 95\% accuracy, significantly higher than any of the individual models, with 90\% of the test samples classified.
\end{abstract}


\section{Introduction}

Enabled by massive parallel computing offered by graphics processing units,
successfully trained deep neural networks have created quantum leaps in performance in
challenging applications~\cite{lecun+al-2015-nature,EfficientPrO2017Sze}, including object recognition, speech recognition, and
natural language processing. Such methods have set the new state of the art records and surpassed human
performance on large datasets. Their applications have created impressive commercial successes,
which have further accelerated the standardization and deployment of deep learning techniques
by developing publicly available frameworks for deep learning. Deep learning techniques are positioned
to change many professions in fundamental ways by offering  decision making capacities more accurate
than human experts (e.g.,~\cite{AFutureThat2017Report}). 


However, the potentials of deep learning techniques are hindered by the lack of
understanding of the very core set of the algorithms and techniques that have enabled their successes.
In the 2017 Test of Time Award Presentation at the NIPS Conference, Rahimi\footnotemark\footnotetext{Video is available at https://www.youtube.com/watch?v=Qi1Yry33TQE.}
compared deep learning to alchemy to bring the attention to
the urgency of developing fundamental understandings.
The successes of deep learning are not consistent with existing statistical learning theory,
which states that generalization of models is a function of their capacity. The larger the capacity of a model,
the worse it should generalize. On the other hand, deep neural networks are typically overparameterized (i.e., they
have more parameters than the number of the training samples) and yet they give best generalization performance
on large datasets using simple optimization techniques like the stochastic gradient descent algorithm.
Given that the loss functions of deep neural networks are highly nonlinear and not convex~\cite{DeepLearningWPoor2016Kawaguchi,QualChar2015GoodF}, 
the only plausible explanation is that the optimization problem for deep neural networks is made easier by having more parameters
and therefore increasing potentially good solutions exponentially. In addition, since the stochastic gradient descent algorithm is effective, analyzing the training dynamics should provide insights about how samples are being processed.

In this paper, we first provide empirical evidence that good solutions are abundant for deep neural networks.
We then examine overfitting, one of the most fundamental problems in deep learning by analyzing the training dynamics.
While overfitting has been widely recognized, almost all the applications focus on avoiding overfitting
by having more data or adjusting neural network architectures or hyperparameters. 
We are able to explain typical overfitting behaviors observed in deep neural networks as the continued gradient updating.
Furthermore, we develop a consensus-based classification algorithm that allows us to avoid overfitting even when the number of
training samples is relatively small. Systematic experimental results demonstrate significant improvements over existing methods.

The paper is organized as follows. In Section II, we present a primal about solution abundance in overparameterized neural networks. The overfitting mechanism for classification is explained in Section III. In Section IV, we describe the proposed consensus-based classification algorithm and illustrate how our method can be used to identify consistently classified samples. Section V illustrates the effectiveness of the proposed method
with experiments. Section VI gives a brief discussion about the assumptions and possible extensions along with the relationship to other existing methods, while Section VII concludes the paper with a brief summary and future work.


\section{Abundance of ``Good" Deep Neural Network Solutions}
Abstractly, a neural network is a parametrized function $f(x;\theta)$, where $x$ is the input, and $\theta$ is a vector that
includes all the parameters (weights and biases). Given a dataset and a loss function, an optimization algorithm
finds a point in the parameter space as a solution. If we limit to local and iterative optimization algorithms like
stochastic gradient descent and its variants, the particular solution depends on the initial parameter values, optimization dynamics, and other random factors such as the order of batches for gradient estimation.

\begin{figure*}[ht]
  \centering
  \subfloat[]{\includegraphics[width=0.450\textwidth]{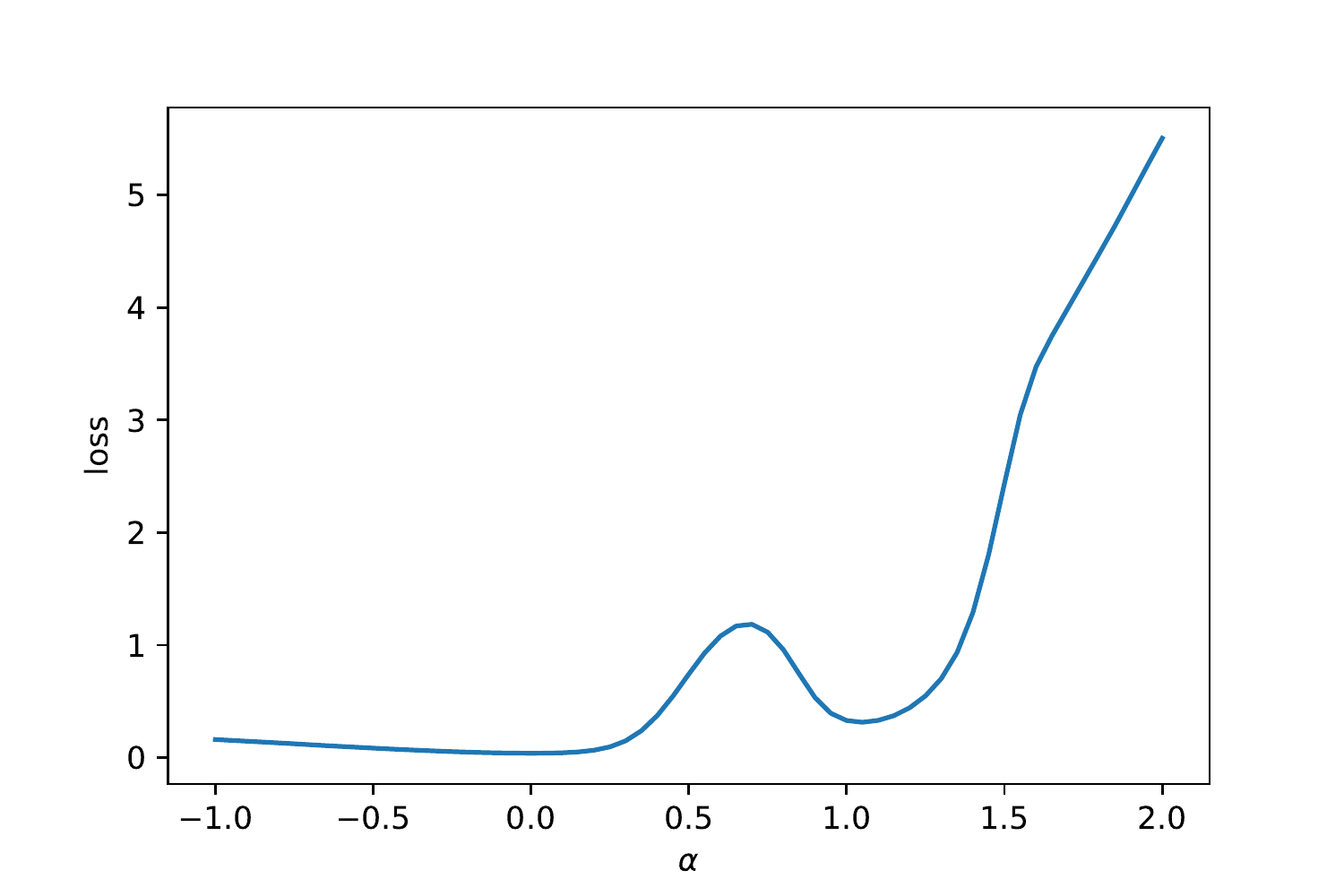}\label{fig:la}}
  \subfloat[]{\includegraphics[width=0.450\textwidth]{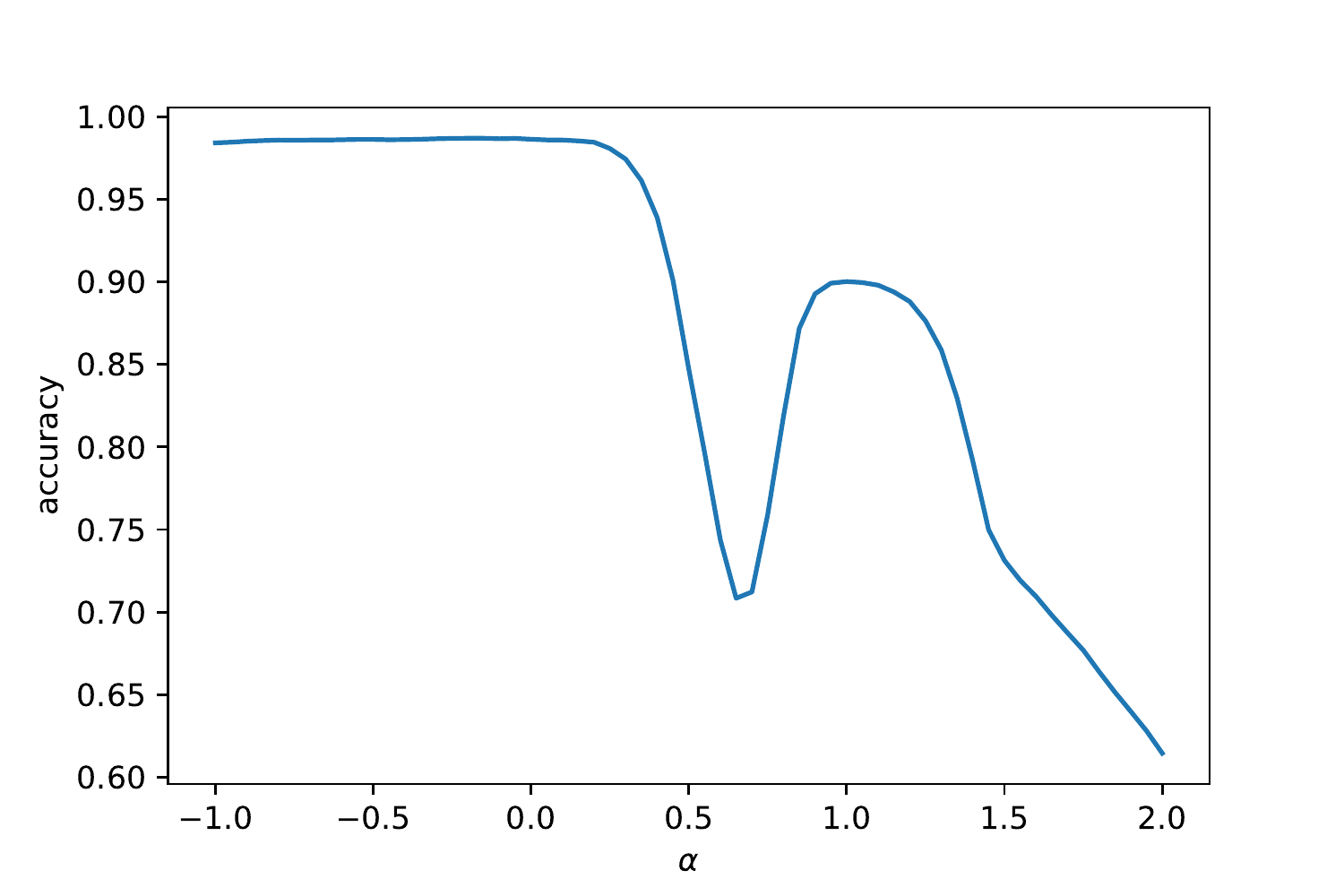}\label{fig:aa}}
  \caption{Good solutions can be reached from any random initialization. Here the model has the parameters that are linearly interpolated from two good parameters based on varying values of $\alpha$. (a) shows the test loss of the model. (b) shows the test accuracy of the model.}
  \label{fig:solution}
\end{figure*}

Even though deep neural networks are overparameterized, still simple optimization methods such as stochastic gradient descent, implemented as backpropagation, can obtain good solutions. This would not be possible if a good solution could not be reached with high probability from a typical initial solution. 
While some prior studies theoretically or empirically show that local minima may not be a significant problem for non-convex optimization of neural networks~\cite{OnTheSaddle2014Pascanu,IdentSaddle2014Dauphin,ExactSol2013Saxe,LossSur2015Chorom}, however, some of those works state that proliferation of saddle points can be a bottleneck. But, Goodfellow et al.~\cite{QualChar2015GoodF} demonstrated that they did not find any evidence of local minima and saddle points slowing down the stochastic gradient descent learning. By interpolating from initial solution to a network solution, they show that there often exists a smooth path in the objective function landscape. 

For further evidence, here we study interpolation between different solutions. More specifically, we interpolate between two specific solutions. To get these two solutions, we first train a simple deep neural network having three layers. The first layer is a Convolutional layer with 32 filters and filter size of 3$\times$3, followed by max pooling of size of 2$\times$2, ReLU activation and Dropout with parameter 0.25. The second layer is a Fully connected layer with 128 units, followed by ReLU activation and Dropout with parameter 0.50. The third layer is a Fully connected softmax layer with 10 units. We trained the network several times from different initial solutions on MNIST dataset~\cite{Mnist1998Lecun} to obtain multiple solutions. We use
two good solutions  (i.e., $w_1$ and $w_2$) and evaluate the loss and accuracy using the interpolated solutions $w = \alpha w_1 + (1-\alpha) w_2$ for varying values of $\alpha$.
Figure \ref{fig:solution} shows that within certain range (-1.0 to 0.4) of $\alpha$, there are many good solutions. The results are further consistent with some recent works that demonstrate that large amount of good solutions are the main reason behind the successes for optimizing deep neural networks. For example, Allen-Zhu et al.~\cite{GoingBeyondTwo} show that good solutions can be reached from almost everywhere. Furthermore, Wu et al.~\cite{Landscape2017TowardsUG} prove
that the volume of basin of attraction of good solutions is much larger compared to that of bad solutions, which generally  helps the optimization methods to get to good solutions.   

\section{Overfitting Mechanism}
The abundance of good solutions seems to be incompatible with overfitting solutions that have been observed
when training deep neural networks. 
After training a deep neural network model on known labeled data, generally it is tested on unseen data to check if it generalizes well.  The model having good generalization capability means that it has the capability of performing well on test data. Overfitting occurs when the model does well with training data and fails to perform well on test data. More specifically, the model learns the noise patterns present in the training data, hence a large gap between the training and test error is seen due to overfitting. In contrast, underfitting happens if the model fails to capture the patterns both in training and test data.
Overfitting is a fundamental issue in machine learning and in deep learning in particular. Deep neural networks are prone to overfitting because of the large number of parameters to be learned. Moreover, these networks are so flexible 
and overparameterized that they adjust the parameters in order to fit the training data;
as demonstrated by~\cite{Zhang2016UnderstandingDL,AcloserLook2017Krueger}, they can perfectly fit image data even with 
labels randomized. 

\begin{figure*}[ht]
  \centering
  \subfloat[]{\includegraphics[width=0.450\textwidth]{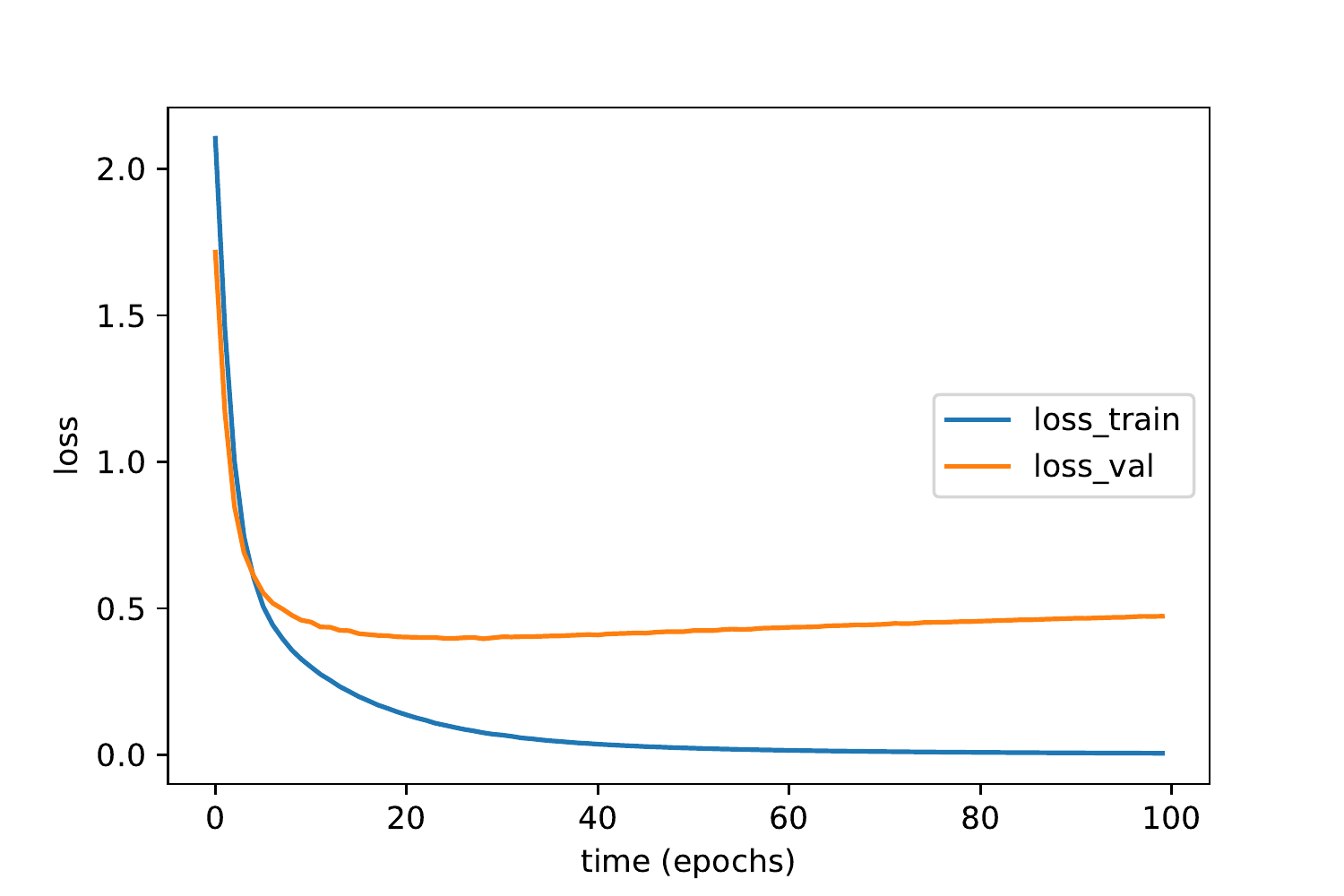}\label{fig:ola}}
  \subfloat[]{\includegraphics[width=0.450\textwidth]{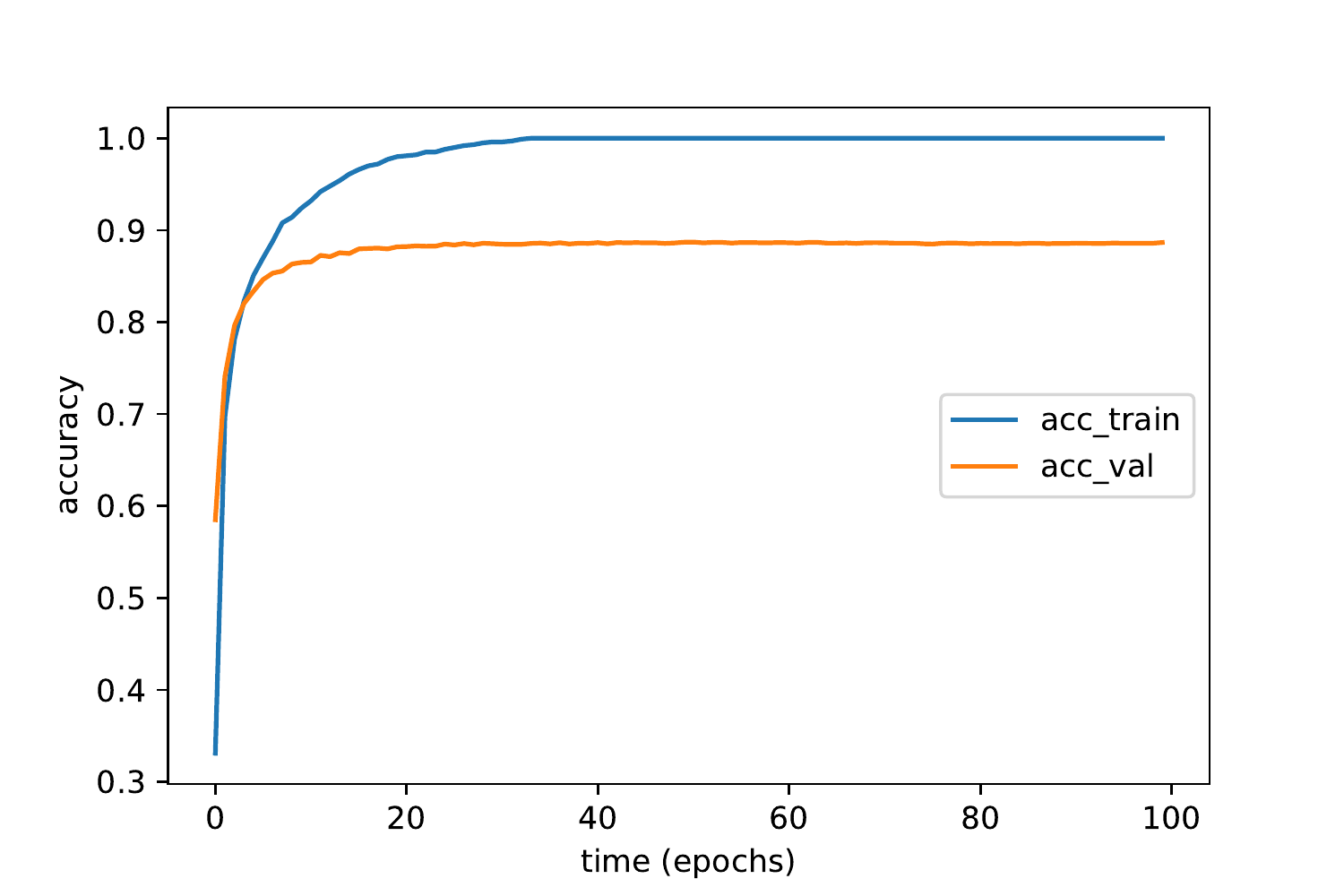}\label{fig:oaa}}
  \caption{Overfitting of the model on MNIST dataset. (a) shows that after a certain epoch, the training loss gets decreased while validation loss gets increased. (b) shows the gap between training and validation accuracy over the time.}
  \label{fig:overfitting}
\end{figure*}

\begin{figure}[ht]
    \centering
    \includegraphics[width=8cm]{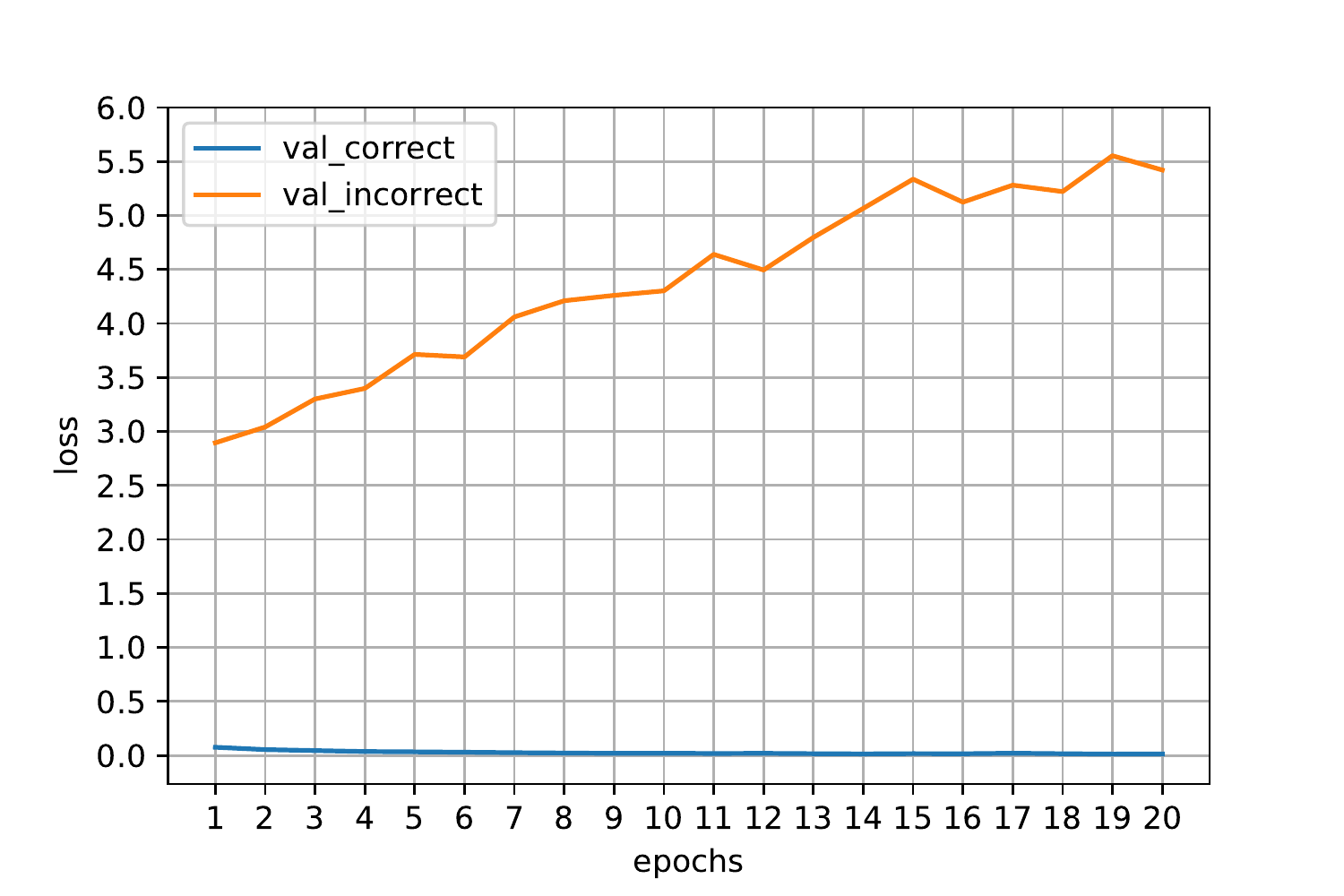}
    \caption{Validation loss for correctly and incorrectly classified samples. Loss is going up for incorrectly classified samples and going down for correctly classified samples.}
    \label{fig:val_loss}
\end{figure}

To illustrate the typical behaviour, we have trained a model on part of the MNIST dataset.
Fig. \ref{fig:overfitting}(a) shows an example that shares the typical overfitting characteristics.  
Here the model is overfitting as the loss on the training data is decreasing but the loss on the validation set
is increasing. 
We show that the increase in loss on the validation set is due to continuous update of the
weights and biases, causing the magnitudes of the inputs to the softmax layer to increase. 
Note that the cross entropy loss used by the softmax layer is sensitive to scaling of its input.
If we scale the input by a factor larger than 1, the probability distribution from the softmax layer will
become more peaked. As a toy example, suppose the original input to the softmax layer is \{0, 0.5, 1, 2\},
the probabilities for the classes are \{0.0784,  0.1292,  0.2131, 0.5793\} respectively. Now if we simply multiply
all the numbers by 1.25, the probabilities for the classes become \{0.0539,  0.1008,  0.1882, 
0.6571\} respectively. Note that the cross entropy loss depends on the correct label for the class.
If the sample belongs to class 4, the loss decreases from 0.5460 to 0.4200.
If the sample belongs to class 1, then the loss increases from 2.5460 to 2.9200.
The changes in loss in Fig. \ref{fig:overfitting}(a) are due to the changes to the magnitudes
of inputs to the softmax layer.
To further illustrate this, Fig. \ref{fig:val_loss} separates the loss on the validate set into
a correctly classified subset and an incorrectly classified subset.
It is clear from the curves that the increase in loss is due to the increase of the incorrectly classified subset;
the loss of the correctly classified subset decreases.

We have verified the observed phenomenon on numerous different examples and using different
models. It is clear that overfitting is not a special phenomenon and it simply indicates that
gradients are not zero and the weights are being updated. This has been observed by the others. For example, Goodfellow et al.~\cite{Goodfellow-et-al-2016} show that gradient increases with respect to training iterations, even after the training performance stabilizes. 
As shown in Fig. \ref{fig:overfitting}(b), the accuracy on the validation set remains almost constant,
further validating the loss is due to the increase of input magnitudes to the softmax layer.



\section{Consensus-based Classification}
While we are able to explain how overfitting happens,
we have not explained the generalization gap between the performance on the training set 
and the one on the validation set as shown in Fig. \ref{fig:overfitting}(b).
To address this issue, we first analyze the training dynamics of samples and then propose 
a consensus-based classification algorithm to overcome the gap.

\subsection{Training Dynamics}

\begin{figure*}[ht]
  \centering
  \subfloat[]{\includegraphics[width=0.3\textwidth]{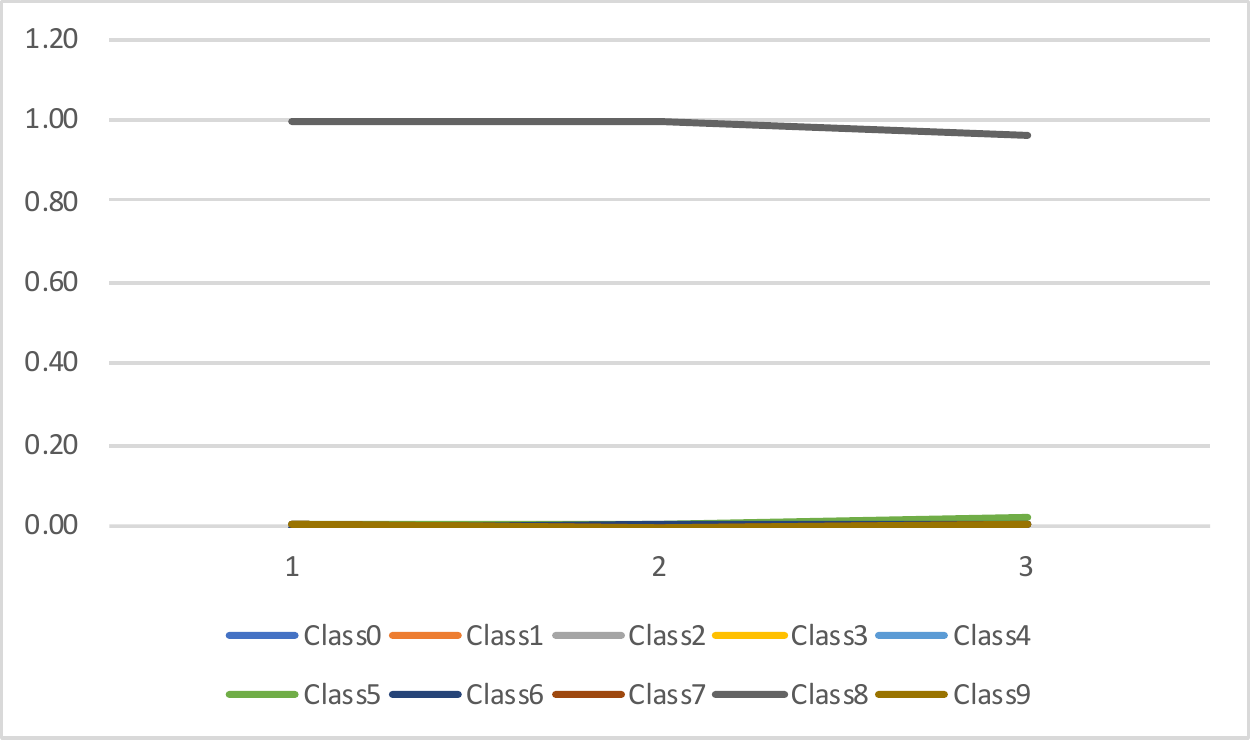}\label{fig:tdg}}
  \subfloat[]{\includegraphics[width=0.3\textwidth]{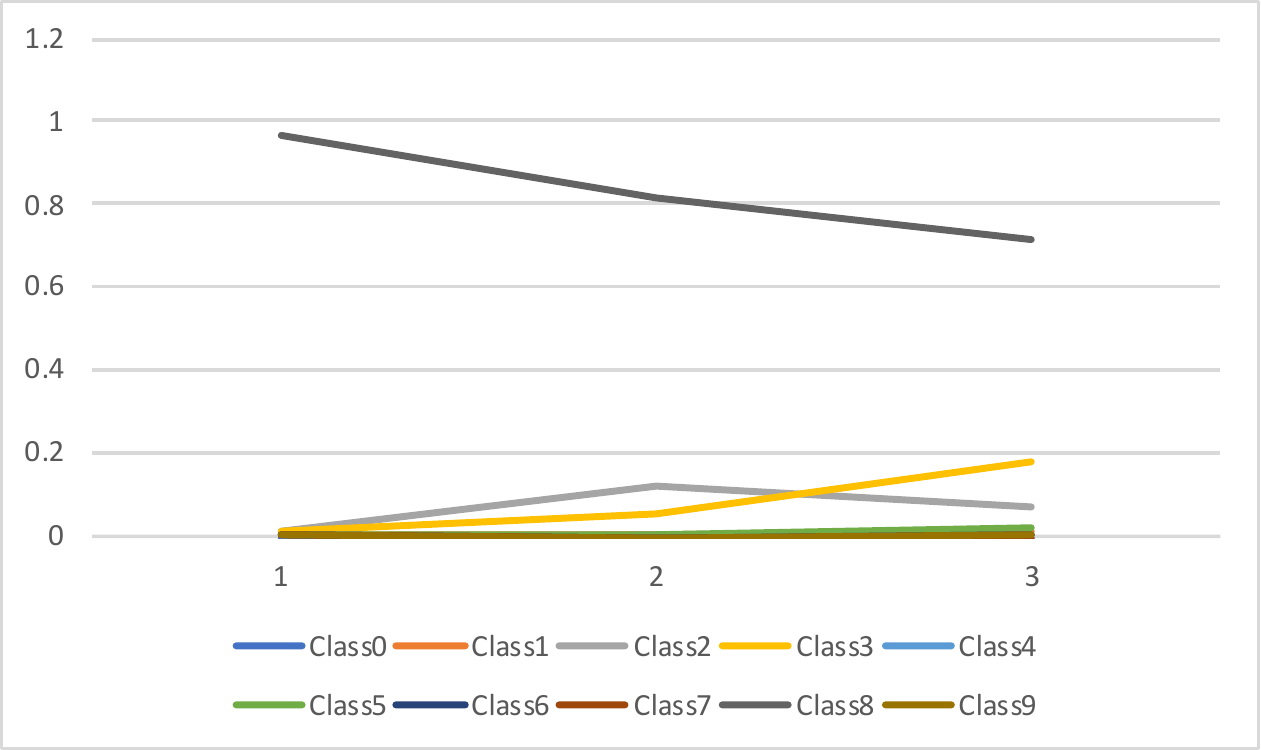}\label{fig:tdg1}}
  \subfloat[]{\includegraphics[width=0.3\textwidth]{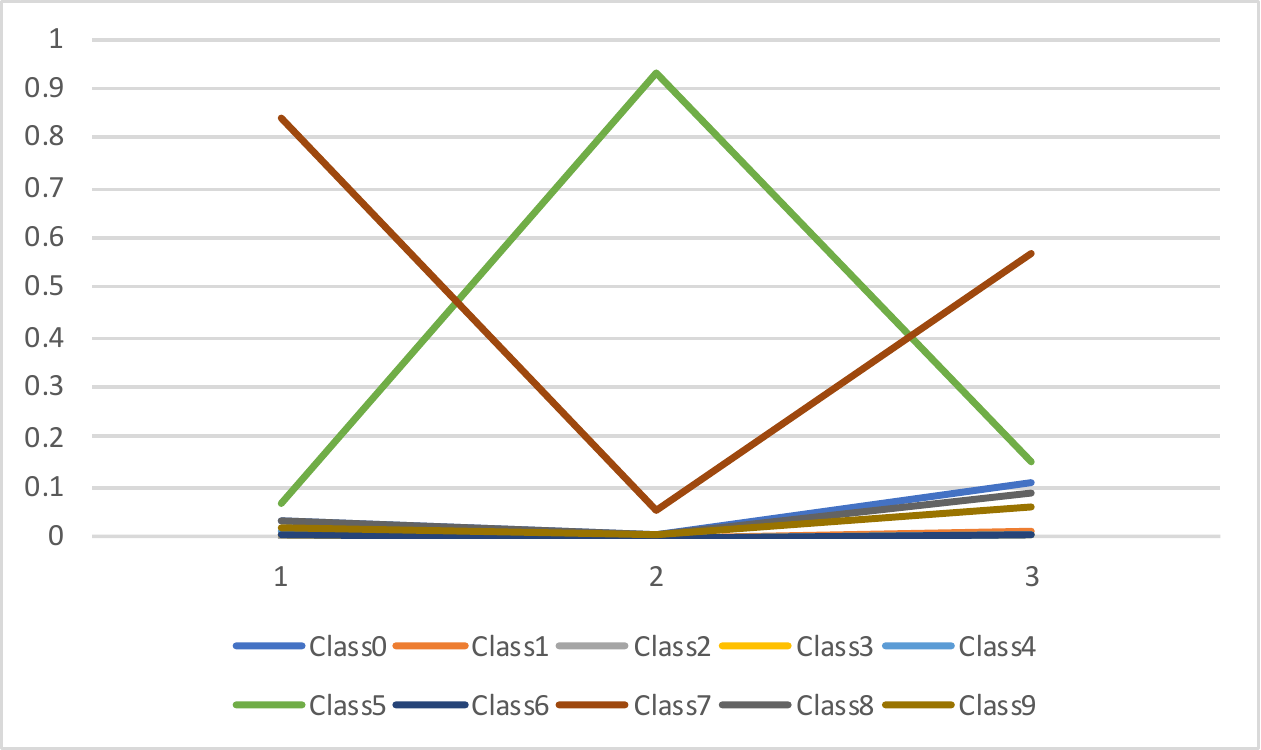}\label{fig:tdb}}
  \caption{Training dynamics of the samples. (a) and (b) show two samples that classified consistently in all three models. (c) shows a sample whose classification is unstable over the models.}
  \label{fig:dynamics}
\end{figure*}
We analyze the training dynamics for each sample and able to get some useful information about the samples. To analyze the training dynamics of the samples, we consider three models having different architectures trained on MNIST dataset. The models are trained on $1k$ training samples while validated on $10k$ validation samples. Figure \ref{fig:dynamics} illustrates the training dynamics of the validation samples. Here we consider three representative samples (e.g., good and bad). Fig. \ref{fig:dynamics}(a) is an example of a good sample. We consider it as a good sample since all the models classify that sample consistently. Because of the stable classification, we can say the models are much more confident about that classification of the sample. Fig. \ref{fig:dynamics}(b) is also similar like \ref{fig:dynamics}(a) except that the prediction from third model is less reliable.  On the other hand, Fig. \ref{fig:dynamics}(c) is an example of a bad sample. It is bad as for this sample, the classification is not stable among the models. As the classification is not stable, it cannot be said for sure what is really going on. The classification is changing, sometimes it is classified as one class in one model while another model classifies it to a totally different class. We conjecture that this unstable classification is due to some random factors, rather than inherent factors supported by the data. Therefore, if we want to classify this sample, we cannot classify with good confidence as sometimes it is classified as one class, sometimes classified as different class. 

\subsection{Consensus-based Classification}
The key observation that motivates the proposed algorithm is that if a sample is classified due to some
random factors by a model, the chance can be reduced exponentially by using multiple models.
The core idea is used for random projections. Rather than choosing the best among all the models,
we should use consensus to identify the samples that are classified consistently and the ones that are classified
randomly.
Here we assume that all the models can capture the intrinsic parts of the samples. On the other hand, the parts that are captured by one model but not by the others, we assume those are extrinsic due to some kind of noise as the probability of getting good solutions in overparameterized network is high~\cite{GoingBeyondTwo}.

The proposed algorithm for classifying any sample is given in Algorithm 1. In general, the algorithm requires $n$ trained models $M_1$,\ldots,$M_n$ and parameter $p_t$, which is a threshold to determine if a model's probability is sufficient. Note that there are different ways to describe and implement the algorithm. To illustrate further the algorithm, we apply it to the three examples shown in Fig. \ref{fig:dynamics}. For any value $p_t < 0.96$, the sample in Fig. \ref{fig:dynamics}(a) will be classified consistently as Class 8.  The sample in Fig. \ref{fig:dynamics}(b) will be classified as Class 8 for any value $p_t < 0.712$; if one uses a value for $p_t$ higher than 0.712, this sample will not be classified by the algorithm.
On the other hand, for the sample in Fig. \ref{fig:dynamics}(c), it will not be classified for any value $p_t \ge 0.1$ as model 2 classifies it differently than model 1 and model 3. In addition, the sample is ambiguous to model 3.

\begin{algorithm}
\caption{Consensus-based Classification}\label{algo1}
\begin{algorithmic}[1]
\REQUIRE Trained models $M_1$, $M_2$,\ldots, $M_n$, input $x$, and parameter $p_t$
\STATE Apply each of the models to classify $x$ and retain the probabilities for each class as $P_{M_i}$
\STATE Compute $P_{min}$ by finding the class-wise minimal among $P_{M_i}$
\STATE If $max(P_{min}) > p_t$,
\STATE \hspace{0.1in}   Classify $x$ as the class with maximum $max(P_{min})$
\STATE Else 
\STATE \hspace{0.1in}    Reject to classify $x$ (mark it as ambiguous)
\STATE Endif
\end{algorithmic}
\end{algorithm}

Our algorithm is different from the ensemble methods. While ensemble methods use the average to improve the performance, we are not using average. In our work, we find the inconsistency as a way to reject the samples that cannot be classified consistently. In contrast, ensemble methods do not reject any sample. In other words, ensemble methods can not avoid over generalization, while our algorithm inherently rejects inputs that may be due to over generalization.

\section{Experimental Results}
We have applied the proposed algorithm on two representative datasets using different deep neural network architectures and optimization
algorithms. We use one type of dataset that contains size normalized handwritten zip code digits scanned from envelopes by the U.S. Postal Service~\cite{HandZipCode1990Lecun}. The dataset consists of 16$\times$16 pixel grayscale images, among them 7291 are training samples and 2007 are test/validation samples. Figures \ref{fig:val_loss} and  \ref{fig:distribution} are based on the experimental results from the models that are trained and validated on this dataset. All the other figures in this paper are based on experiments on the deep learning models trained and validated on the widely used MNIST dataset~\cite{Mnist1998Lecun}. This dataset contains 70,000 handwritten digit images having size 28$\times$28 pixel for each image. Among the images, 60,000 are training samples and 10,000 are validation samples. However, for our experimental purpose we used only 1000 as our training samples out of 60,000. 

\subsection{Density Distribution of the Samples}
\begin{figure*}[ht]
  \centering
  \subfloat[]{\includegraphics[width=0.450\textwidth]{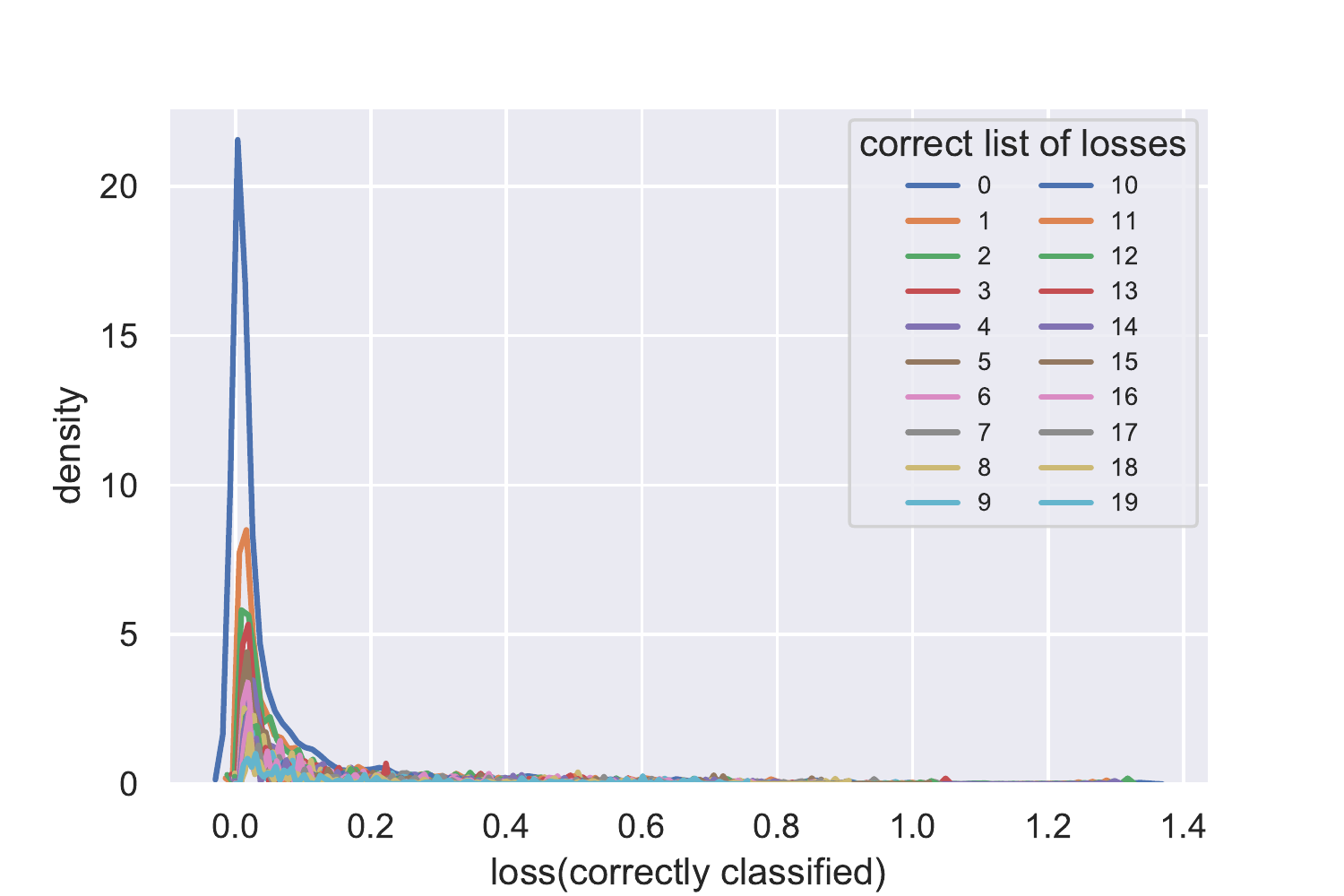}\label{fig:dpc}}
  \subfloat[]{\includegraphics[width=0.450\textwidth]{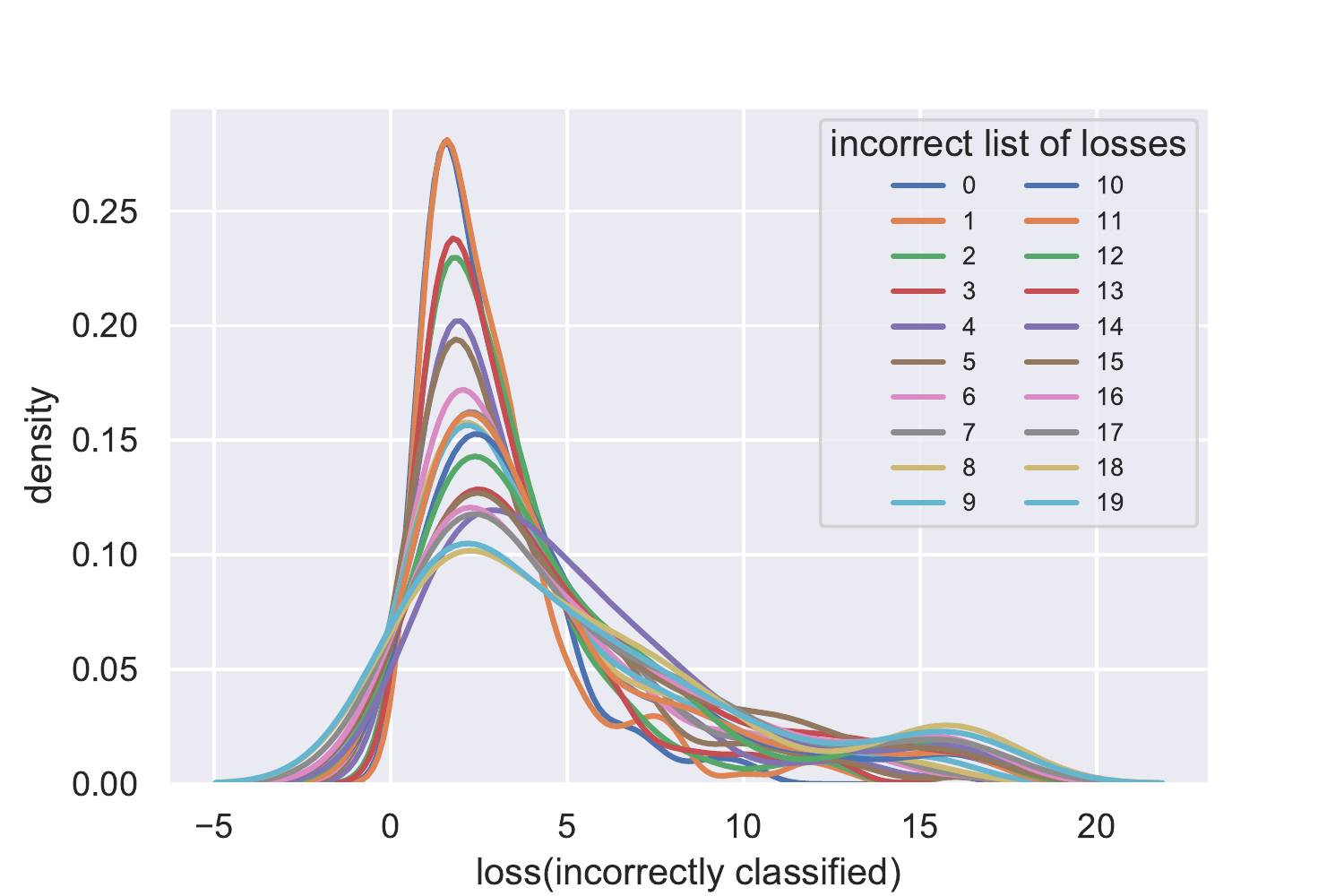}\label{fig:dpi}}
  \caption{Density distribution of the samples over the epochs. (a) shows the distributions of losses for correctly classified samples. (b) shows distributions of losses for incorrectly classified samples.}
  \label{fig:distribution}
\end{figure*}
Here we have tried to identify the overgeneralized samples. For that, we have saved the models during the epochs. We get twenty trained and saved models. After loading the models, for each model, we evaluate the samples in the validation set and get each sample’s individual loss, accuracy and whether they are classified correctly or incorrectly. After that we plot the distributions of the loss values that are classified correctly and the ones that are classified incorrectly. In other words, we get $40$ distributions, $20$ for correctly classified ones and $20$ for incorrectly classified ones. Here, we need to note that a sample may be classified correctly by one model, but may be classified incorrectly by another model. Fig. \ref{fig:distribution} illustrates the distributions for correctly and incorrectly classified samples in each of the model (e.g., epoch). We find that the distributions of losses for correctly and incorrectly classified samples overlap very little. This essentially tells us how the two classes are different as each model show how the loss values are distributed.

\subsection{Consensus-based Classification Using Different Architectures}

\begin{figure*}[ht]
  \centering
  \subfloat[]{\includegraphics[width=0.450\textwidth]{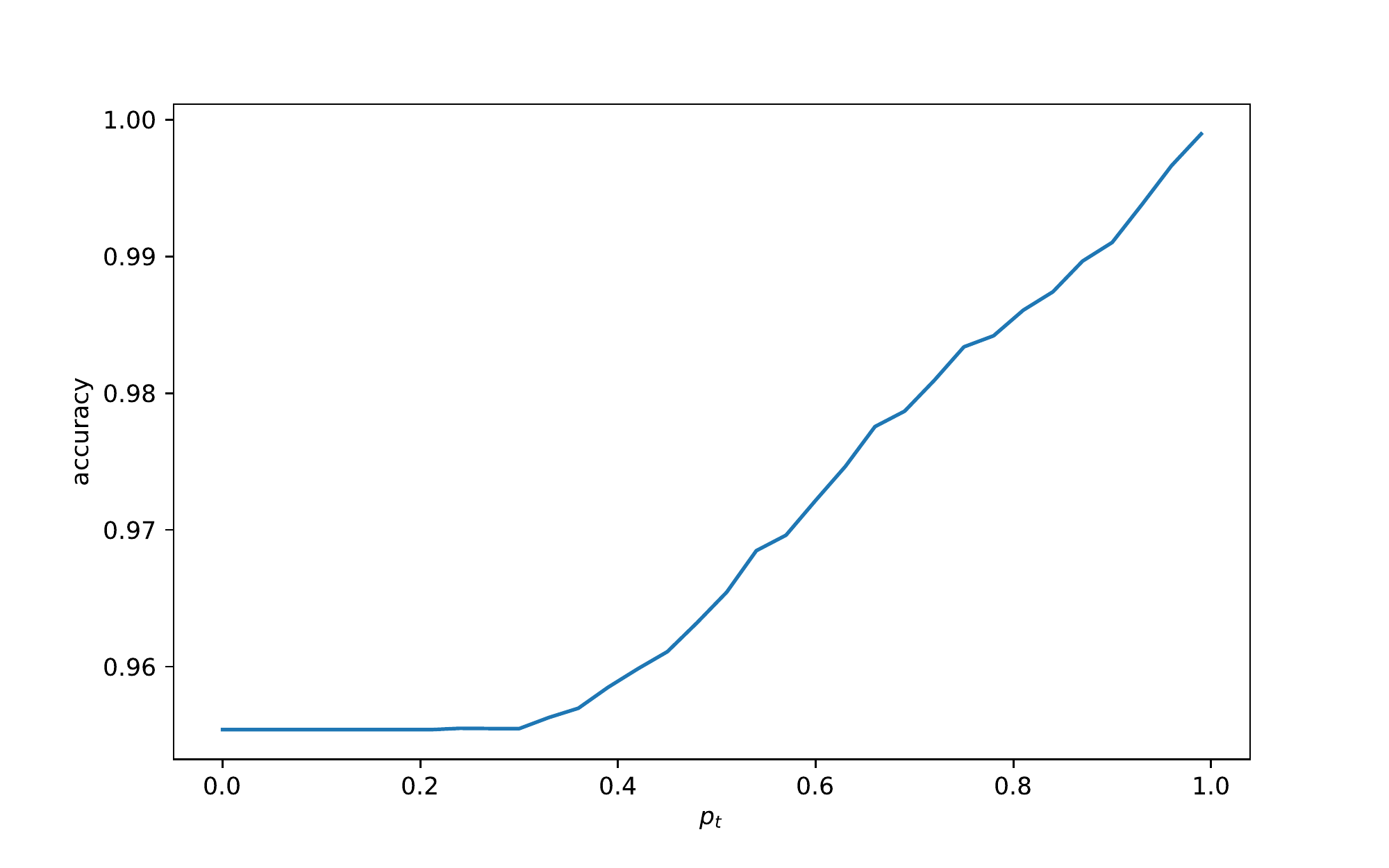}\label{fig:ccs_acc_thres}}
  \subfloat[]{\includegraphics[width=0.450\textwidth]{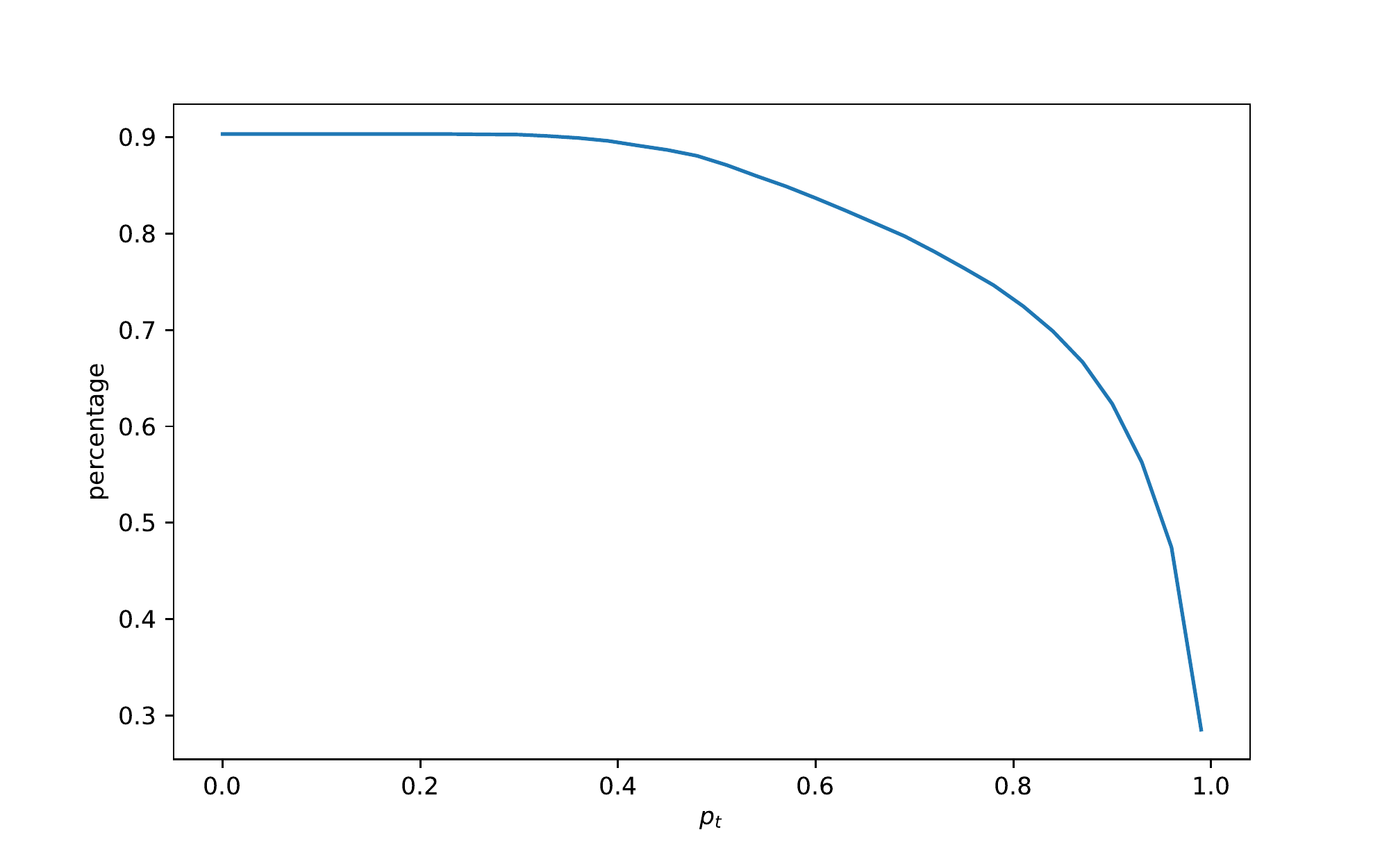}\label{fig:ccs_per_thres}}
  \caption{Change of accuracy and percentage of the CCS samples w.r.t varying threshold values. (a) shows how we can increase the intrinsic accuracy with varying threshold. (b) shows the percentage of CCS samples with varying threshold.
  }
  \label{fig:ccs_acc_per_thres}
\end{figure*}

Here we apply the consensus-based classification on neural networks with different architectures.
We divide the samples into two classes. By considering model dynamics, we classify the samples into extrinsically classified samples and intrinsically classified samples. Intrinsic classification accuracy does not vary significantly as to the dataset size. Because of that reason, we can also term them as consistently classified samples (CCS). On the other hand, samples are extrinsically classified if they get classified due to extrinsic factors such as randomness of weight initialization, softmax outputs, and so on. We believe that randomness factors cannot happen consistently in multiple models. In other words, by examining classification dynamics of multiple models, we can identify samples that are overgeneralized. Therefore we can improve the performance on the remaining samples. Fig. \ref{fig:ccs_acc_per_thres} shows how we are able to improve the intrinsic classification accuracy substantially. Here we only consider one thousand samples for training our three models having different architectures. The extrinsic and intrinsic classification is done on the validation samples. We check that on the three models. A validation sample is intrinsically classified if the probabilities of the three models are higher than the threshold (e.g., $p_t$). Figure \ref{fig:ccs_acc_per_thres} shows that when the threshold is low (e.g., less than 0.4), number of samples that affected by the threshold is very small. Because of that they are giving similar performance when the threshold is low. In contrast, when the threshold is increasing, we see number of correctly classified samples are increasing compared to the total number of consistently classified samples. In other words, by considering the consistently classified samples and by varying the threshold value, we increase the accuracy substantially. As shown by the experimental results, our method improves performance when number of training samples are small. The results are also consistent with recent works~\cite{generalization2018kawaguchi,ModernGenSmall2018Olson}, where the authors argue that generalization is not necessarily dependent on the volume or size of the data.

\begin{figure*}[ht]
\centering
\subfloat[]{\includegraphics[width=0.450\textwidth]{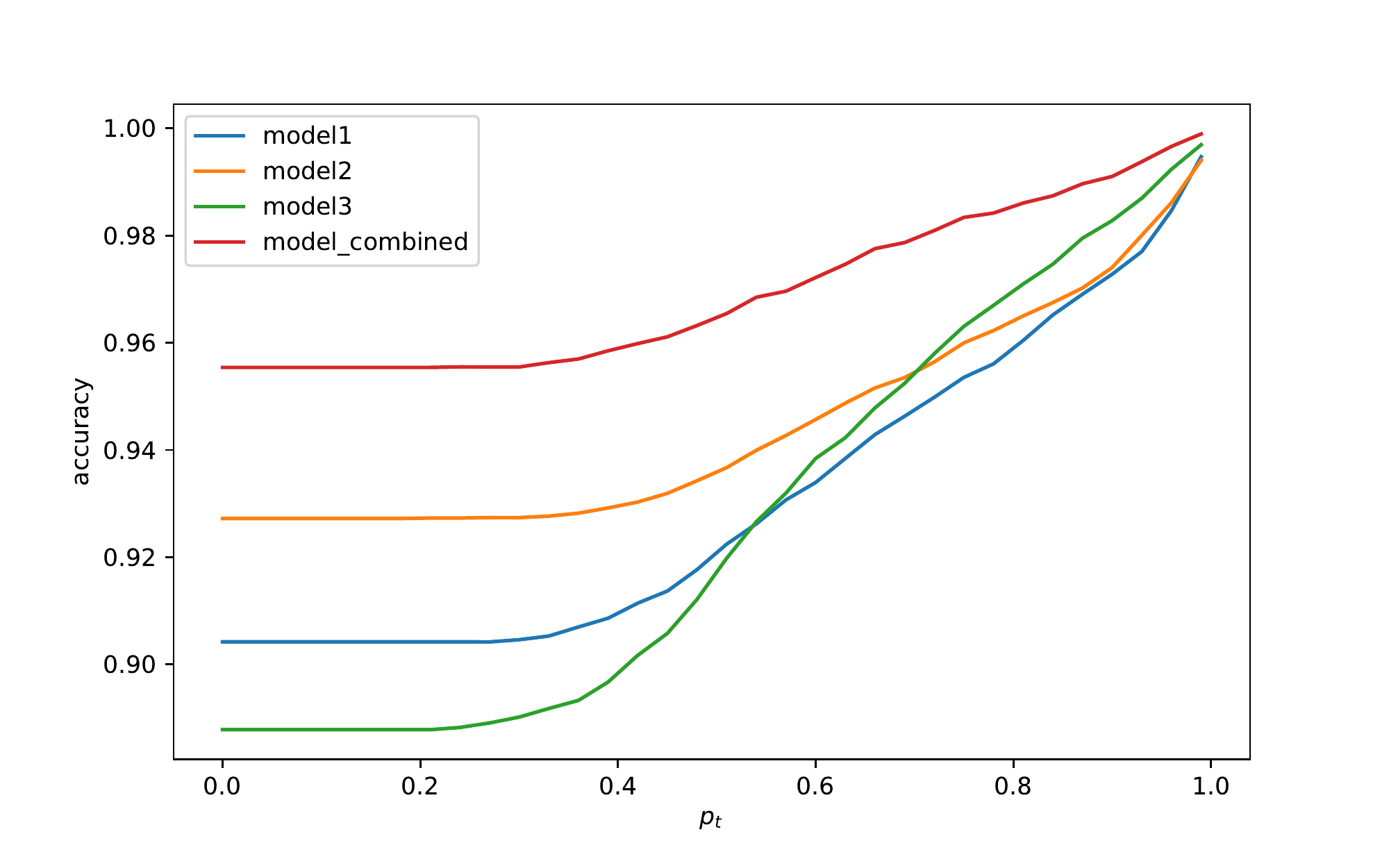}\label{fig:ccs_acc_comb_comp}}
  \subfloat[]{\includegraphics[width=0.450\textwidth]{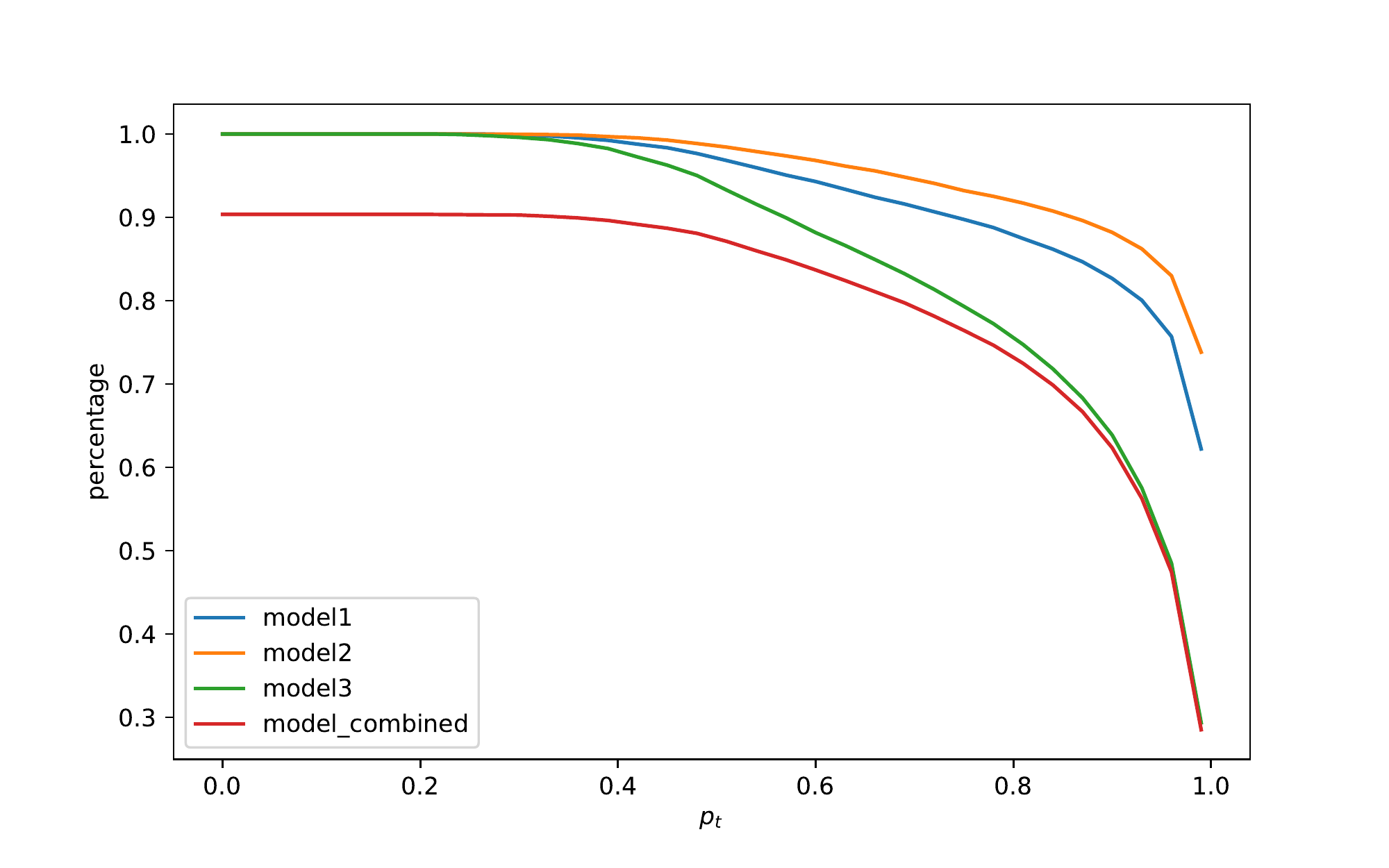}\label{fig:ccs_per_comb_comp}}
\caption{Comparison of results from individual models and the combined model. Here the same algorithm is applied using an individual model and the three combined model. (a) shows the increase of intrinsic accuracy with respect to threshold parameter $p_t$. (b) shows the percentage of the CCS samples for the three models separately and for the combined model.}
\label{fig:ind_comb_comp}
\end{figure*}

To show the effects of using multiple models, Fig. \ref{fig:ind_comb_comp}
shows the results from individual models and also from the consensus model
for comparison.
Note that Algorithm 1 can be applied to a single model as well. In that case,
the threshold is the same as requiring the probability of the correct class to be higher than the given threshold in order for the sample to be classified.
Fig. \ref{fig:ind_comb_comp}(a) shows clearly that the combined model improves the accuracy substantially and consistently for all the values
of threshold. Fig. \ref{fig:ind_comb_comp}(b) shows that the percentage of CCS samples of the three models along with the combined one. Even though model 3 and combined model are similar when $p_t$ is higher than 0.8, they are different. For example, for $p_t=0.99$, there are 2935 CCS samples for model 3 and 2926 of them are classified correctly; for the combined model, there are 2851 CCS samples and 2848 of them are classified correctly.

\subsection{Impact of Dropout in Overfitting}
\begin{figure*}[ht]
  \centering
  \subfloat[]{\includegraphics[width=0.450\textwidth]{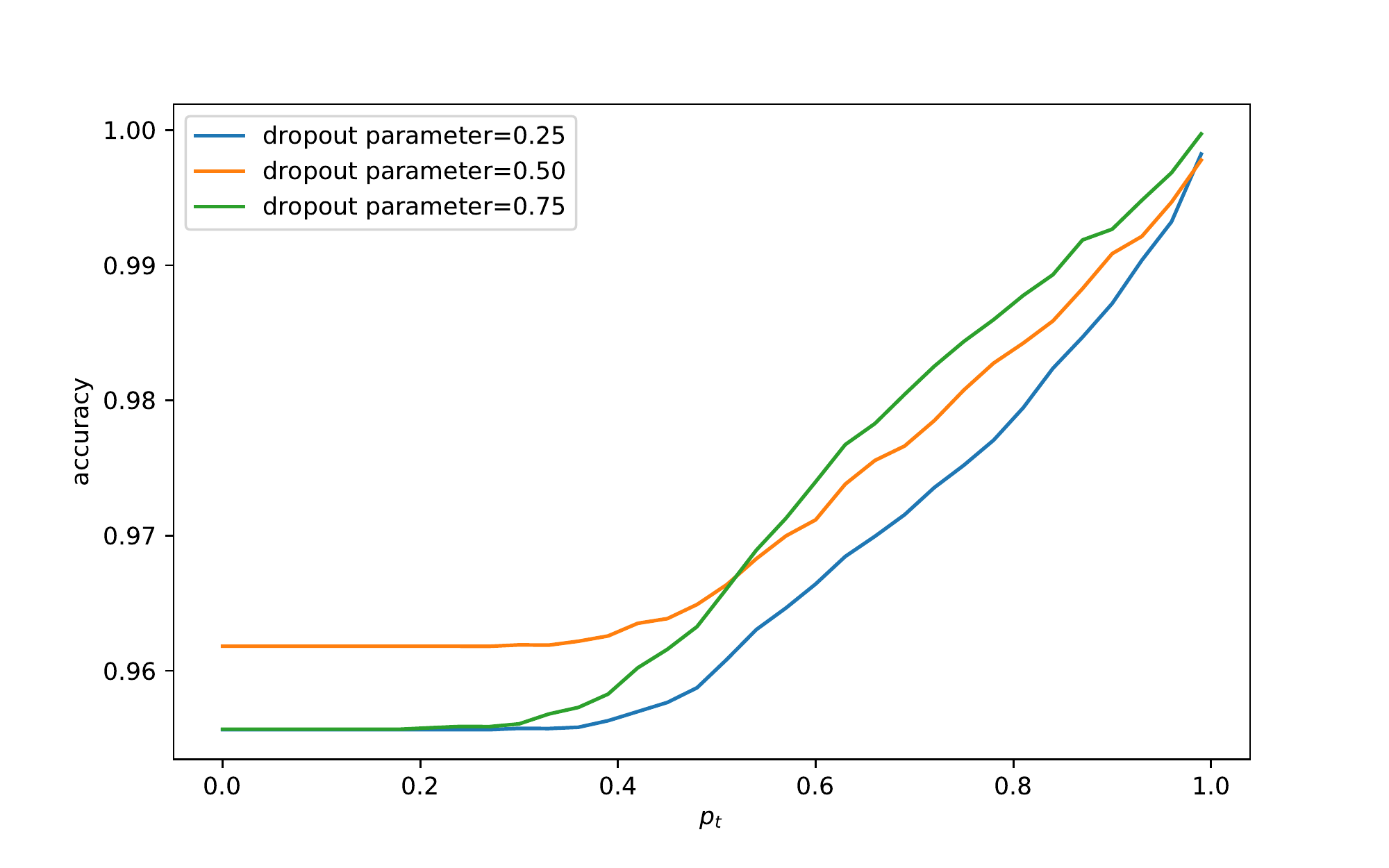}\label{fig:ccs_acc_drop_thres1}}
  \subfloat[]{\includegraphics[width=0.450\textwidth]{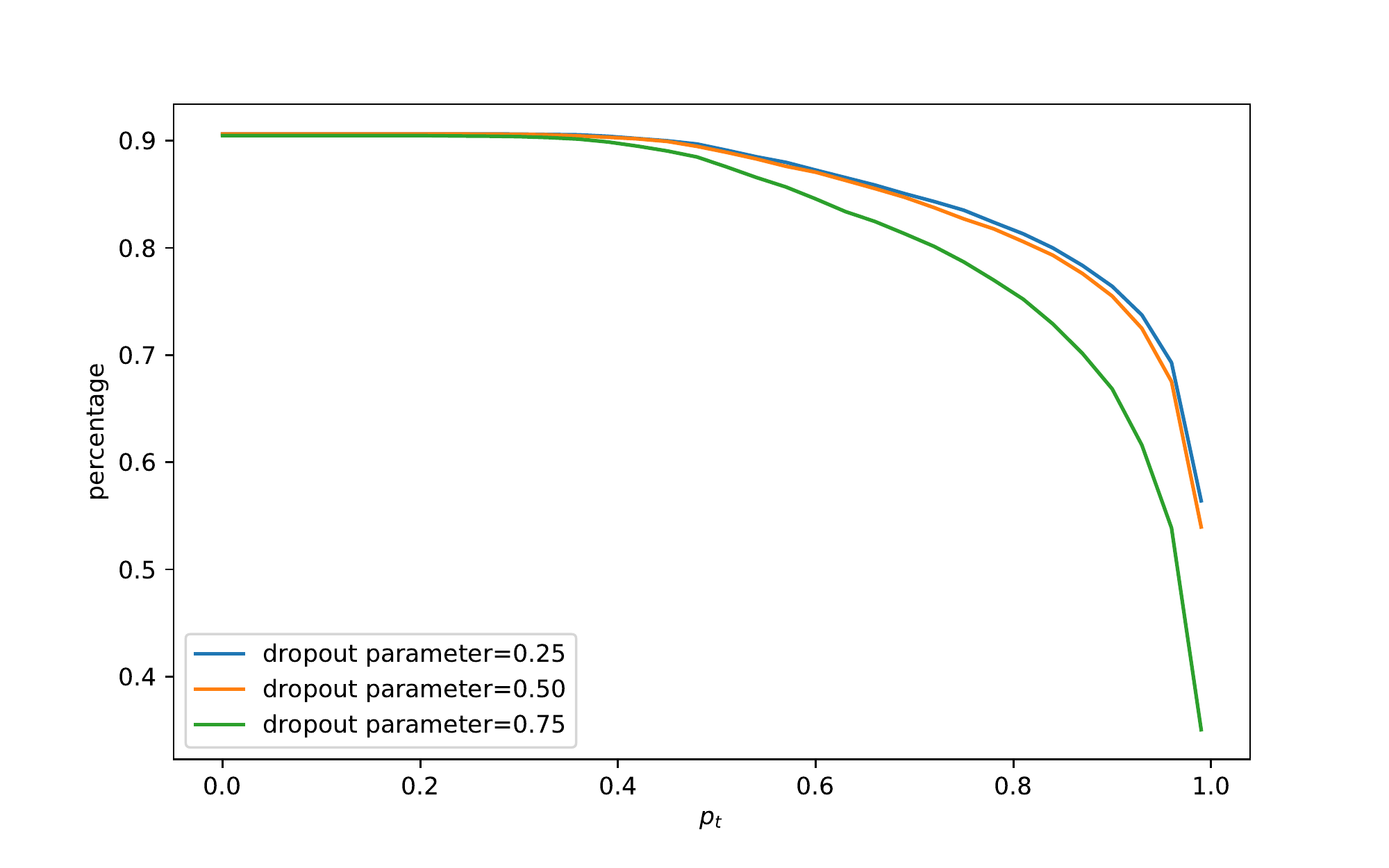}\label{fig:ccs_per_drop_thres2}}
  \caption{Change of accuracy and percentage of the CCS samples w.r.t varying threshold values when dropout used. (a) shows the increase of intrinsic accuracy when three different dropout parameters are used. (b) shows the percentage of the CCS samples for three dropout parameters.}
  \label{fig:ccs_acc_per_drop_thres}
\end{figure*}

Dropout is one of the regularization techniques that is used widely to reduce overfitting in deep neural networks~\cite{dropoutSrivastava14a,Ba2013AdaptiveDF}. It is believed that dropout prevents complex co-adaptations among the neural units. It essentially ignores some of the units randomly during the training phase of the network. In order to work with dropout, a hyperparameter ($p$) needs to be tuned which means that each unit will be retrained with probability $p$, or dropped out with probability $1-p$. 

While dropout somewhat reduces overfitting, however, we believe that dropout does not necessarily prevent overfitting. In overparameterized network, Allen-Zhu et al.~\cite{GoingBeyondTwo} show that good solutions are dense in the parameter space. The regularization techniques merely lead the system to converge to comparatively better solutions among those good solutions~\cite{Landscape2017TowardsUG}. This case is claimed as true for both the implicit regularizer (e.g. SGD)~\cite{Zhang2016UnderstandingDL,TrainFaster2015article} and explicit regularizer (e.g. dropout, weight decay, max-norm)~\cite{Landscape2017TowardsUG}.

We have experimented with three dropout values (0.25, 0.50, 0.75) to see the effect of dropout hyperparameter on the accuracy of consistently classified samples. Figure \ref{fig:ccs_acc_per_drop_thres} shows the result of accuracy and percentage when dropout used. When the threshold is low, we expect dropout parameter (0.50) is turned out to be better than the other two cases. We also observe that the difference between these three are not significant. Furthermore, with dropout, it seems there is not much difference in the curves if you compare the result with Fig. \ref{fig:ccs_acc_per_thres}.

\section{Discussion}
We present an empirical approach which implicitly assume all the neural network models are good. We assume that all the models can capture the intrinsic parts of the samples. The parts that are captured by one model but not by the others, we assume those are extrinsic due to some kind of noise. However, we can generalize it by considering $k$ models to be consistent on the samples among the $n$ models. In our case, we assume $n$ = $k$. Another important aspect of our work is that our algorithm is different than the ensemble methods.

Understanding the underlying mechanisms for the success of deep neural networks is imperative as they are being deployed
to solve problems that are critical to human health and social safety. Danger of relying on core components we do not understand fully is real. For example, Ioffe and Szegedy~\cite{BatchNorm2015Ioffe} point that batch normalization reduces a phenomenon termed as internal covariate shift (ICS) that speeds up training and improves performance. However, a recent paper~\cite{HowBatchNorm2018Santurkar} argues that batch normalization actually smooths the optimization landscape and points out the reasons given by the original authors~\cite{BatchNorm2015Ioffe} are tenuous at least.


One of the fundamental problems in deep learning is generalization. Without understanding the mechanisms how generalization works
for deep neural networks, using and experimenting with different choices of architectures, optimization algorithms,
and hyperparameters shares certain attributes of alchemy. In order to move forward to chemistry, where different interaction patterns are documents, we need to systematically analyze different components and also the interactions among the different components.
To reach quantum chemistry for deep learning, we need to understand exactly how different components behavior. Note that
ultimate resolution for deep neural networks is the individual neurons, quantum chemistry style for deep learning is certainly
achievable.

One of the obstacles to achieve trustworthy deep neural network based models
is the existences of adversarial examples. Note that adversarial examples are not ``adversarial" and they are inputs that are very close to decision boundaries. Because of that, unnoticeable changes to human users can
cause the class label to change from one to other. Such examples 
present a challenge when the underlying models need to be robust and trustworthy.
One advantage of the proposed consensus-based classification algorithm is that it is much more difficult to
have adversarial examples to all the models at the same time. By having
more models, the probability of having adversarial examples decreases
exponentially under the assumption that the models are independent. The effects of the proposed algorithm on real datasets will be quantified in further studies.

Note that generalization is a well defined problem when the underlying class-conditional and prior probability distributions
are known. However, they are not useful for deep neural networks.
It is often stated that neural networks behave like nearest neighbor algorithms. Typical overfitting curves show
that accuracy improvement on the training set does not improve after some iterations and this is not consistent with nearest neighbor algorithms. As more training samples are classified correctly, their neighbors should be classified more accurately.

\section{Conclusion and Future Work} 

We have proposed a consensus-based overfitting avoidance algorithm
that allows us to identify samples that are classified due to random factors using multiple models. Our idea is novel in the sense that we have showed how to avoid overfitting after identifying the overgeneralized samples based on the training dynamics. We believe that our results are significant that might allow us to develop useful algorithms to improve applications of deep neural networks.  

In this paper, we have only considered neural networks for classification using the softmax layer. We expect the same general mechanism would apply to other neural network architectures, which we will investigate further.  The algorithm can also be extended to require consensus among $k$ out of $n$ models. We also like to investigate the connections between the proposed algorithm and the commonly used ensemble algorithms (such as bagging and boosting). In addition, we plan to investigate the solution space by examining solutions in planes in addition to the curves we have studied. Furthermore, we plan to investigate how the proposed algorithm can be used to improve active learning by identifying effective samples for better generalization to larger validation sets. 




%

\def\V{\rm vol.~}
\def\N{no.~}
\def\pp{pp.~}
\def\Pot{\it Proc. }
\def\IJCNN{\it International Joint Conference on Neural Networks\rm }
\def\ACC{\it American Control Conference\rm }
\def\SMC{\it IEEE Trans. Systems\rm , \it Man\rm , and \it Cybernetics\rm }

\def\handb{ \it Handbook of Intelligent Control: Neural\rm , \it
    Fuzzy\rm , \it and Adaptive Approaches \rm }

\bibliographystyle{IEEEtran}
\bibliography{main.bib}

\begin{thebibliography}{10}
\providecommand{\url}[1]{#1}
\csname url@samestyle\endcsname
\providecommand{\newblock}{\relax}
\providecommand{\bibinfo}[2]{#2}
\providecommand{\BIBentrySTDinterwordspacing}{\spaceskip=0pt\relax}
\providecommand{\BIBentryALTinterwordstretchfactor}{4}
\providecommand{\BIBentryALTinterwordspacing}{\spaceskip=\fontdimen2\font plus
\BIBentryALTinterwordstretchfactor\fontdimen3\font minus
  \fontdimen4\font\relax}
\providecommand{\BIBforeignlanguage}[2]{{%
\expandafter\ifx\csname l@#1\endcsname\relax
\typeout{** WARNING: IEEEtran.bst: No hyphenation pattern has been}%
\typeout{** loaded for the language `#1'. Using the pattern for}%
\typeout{** the default language instead.}%
\else
\language=\csname l@#1\endcsname
\fi
#2}}
\providecommand{\BIBdecl}{\relax}
\BIBdecl

\bibitem{lecun+al-2015-nature}
\BIBentryALTinterwordspacing
Y.~LeCun, Y.~Bengio, and G.~Hinton, ``Deep learning,'' \emph{Nature}, vol. 521,
  no. 7553, pp. 436--444, May 2015. [Online]. Available:
  \url{http://dx.doi.org/10.1038/nature14539}
\BIBentrySTDinterwordspacing

\bibitem{EfficientPrO2017Sze}
V.~Sze, Y.-H. Chen, T.-J. Yang, and J.~S. Emer, ``Efficient processing of deep
  neural networks: A tutorial and survey,'' \emph{Proceedings of the IEEE},
  vol. 105, pp. 2295--2329, 2017.

\bibitem{AFutureThat2017Report}
\BIBentryALTinterwordspacing
J.~Manyika, M.~Chui, M.~Miremadi, J.~Bughin, K.~George, P.~Willmott, and
  M.~Dewhurst, ``A future that works: Automation, employment, and
  productivity,'' McKinsey Global Institute, Tech. Rep., 2017. [Online].
  Available:
  \url{https://www.mckinsey.com/~/media/McKinsey/Featured%20Insights/Digital%20Disruption/Harnessing%20automation%20for%20a%20future%20that%20works/MGI-A-future-that-works_Full-report.ashx}
\BIBentrySTDinterwordspacing

\bibitem{DeepLearningWPoor2016Kawaguchi}
K.~Kawaguchi, ``Deep learning without poor local minima,'' in \emph{Advances in
  Neural Information Processing Systems 29}, D.~D. Lee, M.~Sugiyama, U.~V.
  Luxburg, I.~Guyon, and R.~Garnett, Eds., 2016, pp. 586--594.

\bibitem{QualChar2015GoodF}
I.~Goodfellow, O.~Vinyals, and A.~Saxe, ``Qualitatively characterizing neural
  network optimization problems,'' in \emph{International Conference on
  Learning Representations}, 2015.

\bibitem{OnTheSaddle2014Pascanu}
R.~Pascanu, Y.~N. Dauphin, S.~Ganguli, and Y.~Bengio, ``On the saddle point
  problem for non-convex optimization,'' \emph{CoRR}, vol. abs/1405.4604, 2014.

\bibitem{IdentSaddle2014Dauphin}
Y.~N. Dauphin, R.~Pascanu, C.~Gulcehre, K.~Cho, S.~Ganguli, and Y.~Bengio,
  ``Identifying and attacking the saddle point problem in high-dimensional
  non-convex optimization,'' in \emph{Proceedings of the 27th International
  Conference on Neural Information Processing Systems - Volume 2}, ser.
  NIPS'14, 2014, pp. 2933--2941.

\bibitem{ExactSol2013Saxe}
A.~M. Saxe, J.~L. McClelland, and S.~Ganguli, ``Exact solutions to the
  nonlinear dynamics of learning in deep linear neural networks,'' \emph{CoRR},
  vol. abs/1312.6120, 2013.

\bibitem{LossSur2015Chorom}
A.~Choromanska, M.~Henaff, M.~Mathieu, G.~{Ben Arous}, and Y.~LeCun,
  ``\BIBforeignlanguage{English (US)}{The loss surfaces of multilayer
  networks},'' \emph{\BIBforeignlanguage{English (US)}{Journal of Machine
  Learning Research}}, vol.~38, pp. 192--204, 2015.

\bibitem{Mnist1998Lecun}
Y.~Lecun, L.~Bottou, Y.~Bengio, and P.~Haffner, ``Gradient-based learning
  applied to document recognition,'' \emph{Proceedings of the IEEE}, vol.~86,
  no.~11, pp. 2278--2324, Nov 1998.

\bibitem{GoingBeyondTwo}
Z.~Allen-Zhu, Y.~Li, and Y.~Liang, ``Learning and generalization in
  overparameterized neural networks, going beyond two layers,'' \emph{CoRR},
  vol. abs/1811.04918, 2018.

\bibitem{Landscape2017TowardsUG}
L.~Wu, Z.~Zhu, and E.~Weinan, ``Towards understanding generalization of deep
  learning: Perspective of loss landscapes,'' \emph{CoRR}, vol. abs/1706.10239,
  2017.

\bibitem{Zhang2016UnderstandingDL}
C.~Zhang, S.~Bengio, M.~Hardt, B.~Recht, and O.~Vinyals, ``Understanding deep
  learning requires rethinking generalization,'' \emph{CoRR}, vol.
  abs/1611.03530, 2016.

\bibitem{AcloserLook2017Krueger}
D.~Krueger, N.~Ballas, S.~Jastrzebski, D.~Arpit, M.~S. Kanwal, T.~Maharaj,
  E.~Bengio, A.~Fischer, A.~Courville, S.~Lacoste-Julien, and Y.~Bengio, ``A
  closer look at memorization in deep networks,'' in \emph{{ICML}}, 2017,
  arxiv:1706.05394.

\bibitem{Goodfellow-et-al-2016}
I.~Goodfellow, Y.~Bengio, and A.~Courville, \emph{Deep Learning}.\hskip 1em
  plus 0.5em minus 0.4em\relax Cambridge, MA, USA: MIT Press, 2016, p. 276.

\bibitem{HandZipCode1990Lecun}
Y.~L. Cun, O.~Matan, B.~Boser, J.~S. Denker, D.~Henderson, R.~E. Howard,
  W.~Hubbard, L.~D. Jacket, and H.~S. Baird, ``Handwritten zip code recognition
  with multilayer networks,'' in \emph{[1990] Proceedings. 10th International
  Conference on Pattern Recognition}, vol.~ii, June 1990, pp. 35--40 vol.2.

\bibitem{generalization2018kawaguchi}
L.~P.~K. Kenji~Kawaguchi and Y.~Bengio, ``Generalization in deep learning,'' in
  \emph{Mathematics of Deep Learning, Cambridge University Press, in
  preparation. Prepint avaliable as: MIT-CSAIL-TR-2018-014, Massachusetts
  Institute of Technology}, 2018.

\bibitem{ModernGenSmall2018Olson}
M.~Olson, A.~Wyner, and R.~Berk, ``Modern neural networks generalize on small
  data sets,'' in \emph{Advances in Neural Information Processing Systems 31},
  S.~Bengio, H.~Wallach, H.~Larochelle, K.~Grauman, N.~Cesa-Bianchi, and
  R.~Garnett, Eds., 2018, pp. 3623--3632.

\bibitem{dropoutSrivastava14a}
N.~Srivastava, G.~Hinton, A.~Krizhevsky, I.~Sutskever, and R.~Salakhutdinov,
  ``Dropout: A simple way to prevent neural networks from overfitting,''
  \emph{Journal of Machine Learning Research}, vol.~15, pp. 1929--1958, 2014.

\bibitem{Ba2013AdaptiveDF}
J.~Ba and B.~J. Frey, ``Adaptive dropout for training deep neural networks,''
  in \emph{NIPS}, 2013.

\bibitem{TrainFaster2015article}
M.~Hardt, B.~Recht, and Y.~Singer, ``Train faster, generalize better: Stability
  of stochastic gradient descent,'' in \emph{Proceedings of the 33rd
  International Conference on International Conference on Machine Learning -
  Volume 48}, ser. ICML'16, 2016, pp. 1225--1234.

\bibitem{BatchNorm2015Ioffe}
S.~Ioffe and C.~Szegedy, ``Batch normalization: Accelerating deep network
  training by reducing internal covariate shift,'' 2015.

\bibitem{HowBatchNorm2018Santurkar}
S.~Santurkar, D.~Tsipras, A.~Ilyas, and A.~Madry, ``How does batch
  normalization help optimization?'' in \emph{Advances in Neural Information
  Processing Systems 31}, S.~Bengio, H.~Wallach, H.~Larochelle, K.~Grauman,
  N.~Cesa-Bianchi, and R.~Garnett, Eds.\hskip 1em plus 0.5em minus 0.4em\relax
  Curran Associates, Inc., 2018, pp. 2488--2498.

\end{thebibliography}
\end{document}